\documentclass[journal,twoside,web]{ieeecolor}
\usepackage{generic}
\usepackage{cite}
\usepackage{amsmath,amssymb,amsfonts}
\usepackage{algorithmic}
\usepackage{graphicx}
\usepackage{algorithm,algorithmic}
\usepackage{hyperref}
\usepackage{booktabs}
\usepackage{multirow}
\usepackage{bm}
\hypersetup{hidelinks=true}
\usepackage{textcomp}
\def\BibTeX{{\rm B\kern-.05em{\sc i\kern-.025em b}\kern-.08em
    T\kern-.1667em\lower.7ex\hbox{E}\kern-.125emX}}
\begin{document}

\title{A Systematic Evaluation of Sample-Level Tokenization Strategies for MEG Foundation Models}

\author{SungJun Cho, Chetan Gohil, Rukuang Huang, Oiwi Parker Jones, and Mark W. Woolrich
\thanks{This work was supported in part by the Medical Research Council under Grant MR/W006731/1, Grant MR/X00757X/1, and the New Therapeutics in Alzheimer's Disease (NTAD); in part by the Wellcome Trust under Grant 106183/Z/14/Z and Grant 215573/Z/19/Z; in part by the Dementia Platform UK under Grant RG94383 and Grant RG89702; in part by the EPSRC Centre for Doctoral Training in Health Data Science under Grant EP/S02428X/1; in part by the Royal Society under Grant RG/R1/241267; in part by the National Science Foundation under Grant 2314493; in part by the NFRF under Grant NFRFT-2022-00241; in part by the SSHRC under Grant 895-2023-1022; in part by the NIHR Oxford Health Biomedical Research Centre under Grant NIHR203316; in part by the Advanced Research + Invention Agency under Grant SCNI-SE01-P004; in part by the Hertford Claire Clifford Lusardi Scholarship; and in part by the Nuffield Department of Clinical Neurosciences. The views expressed are those of the authors and not necessarily those of the NIHR or the UK Department of Health and Social Care. (Corresponding author: SungJun Cho.)}
\thanks{SungJun Cho, Chetan Gohil, Rukuang Huang, and Mark W. Woolrich are with the Oxford Centre for Human Brain Activity, University of Oxford, Oxford OX3 7JX, U.K. SungJun Cho is also with the Nuffield Department of Clinical Neurosciences, University of Oxford, Oxford OX3 9DU, U.K., and Rukuang Huang and Mark W. Woolrich are also with the Department of Psychiatry, University of Oxford, Oxford OX3 7JX, U.K. (e-mail: sungjun.cho@ndcn.ox.ac.uk; chetan.gohil@psych.ox.ac.uk; rukuang.huang@psych.ox.ac.uk; mark.woolrich@ohba.ox.ac.uk).}
\thanks{Oiwi Parker Jones is with the Oxford Robotics Institute, Department of Engineering Science, University of Oxford, Oxford OX2 6NN, U.K. (e-mail: oiwi.parkerjones@eng.ox.ac.uk).}}

\maketitle

\begin{abstract}
Recent success in natural language processing has motivated growing interest in large-scale foundation models for neuroimaging data. Such models often require discretization of continuous neural time series data, a process referred to as `tokenization'. However, the impact of different tokenization strategies for neural data is currently poorly understood. In this work, we present a systematic evaluation of sample-level tokenization strategies for transformer-based large neuroimaging models (LNMs) applied to magnetoencephalography (MEG) data. We compare learnable and non-learnable tokenizers by examining their signal reconstruction fidelity and their impact on subsequent foundation modeling performance (token prediction, biological plausibility of generated data, preservation of subject-specific information, and performance on downstream tasks). For the learnable tokenizer, we introduce a novel approach based on an autoencoder. Experiments were conducted on three publicly available MEG datasets spanning different acquisition sites, scanners, and experimental paradigms. Our results show that both learnable and non-learnable discretization schemes achieve high reconstruction accuracy and broadly comparable performance across most evaluation criteria, suggesting that simple fixed sample-level tokenization strategies can be used in the development of neural foundation models. The code is available at \url{https://github.com/OHBA-analysis/Cho2026_Tokenizer}.\\
\end{abstract}


\section{Introduction} \label{sec1}
\IEEEPARstart{F}{oundation} models are trained on large-scale data to learn representations that can be efficiently adapted to a wide range of downstream tasks. Their success in natural language processing and computer vision has motivated analogous efforts in neuroimaging, aiming to develop Large Neuroimaging Models (LNMs) that leverage abundant unlabeled neural recordings while minimizing reliance on scarce task-specific or clinical annotations.

Learning transferable representations of brain activity using LNMs has the potential to support various neuroscientific and clinical applications, including neural decoding \cite{Thomas_2022, Yao_2025}, biomarker identification \cite{Craik_2019, Rafsani_2025}, and brain-computer interfaces \cite{Tang_2023, Fang_2025, Wang_2025a}. This paradigm is particularly well suited to electrophysiological modalities such as electroencephalography (EEG) and magnetoencephalography (MEG), which provide high temporal resolution and yield large-scale multivariate time-series data \cite{Sejnowski_2014, Stokes_2015}.

In recent years, several foundation models for EEG and MEG have been proposed, predominantly based on the transformer architecture. These models largely fall into three classes: encoder-only transformers trained via masked token prediction (e.g., LaBraM \cite{Jiang_2024}, CBraMod \cite{Wang_2025b}, BrainOmni \cite{Xiao_2025}); encoder-decoder masked autoencoders trained via masked token reconstruction (e.g., REVE \cite{Ouahidi_2025}); and decoder-only autoregressive models trained via next-token prediction (e.g., Neuro-GPT \cite{Cui_2024}, MEG-GPT \cite{Huang_2025}).

Despite this progress, a central but underexplored design choice made in transformer-based neural foundation models is the tokenization: the process of converting continuous time series data to discrete `tokens' \cite{Vaswani_2017}. This choice determines the representational granularity of the data and can introduce an inductive bias. Inappropriate tokenization may obscure biologically meaningful structure or impose assumptions misaligned with the statistical properties of neural data, ultimately limiting representational fidelity and downstream performance. In this sense, the tokenization is not simply a preprocessing step but a defining component that can determine the success of a neural foundation model. An effective tokenizer must therefore encode neural dynamics in a form that preserves the temporal and spectral structure while remaining computationally tractable.

Existing tokenization strategies for M/EEG LNMs have largely been drawn from general-purpose time-series modeling and can be broadly categorized based on the temporal resolution of the resulting tokens. \textit{Sample-level} tokenizers map each time point to a token, preserving native temporal resolution and spectral content, whereas \textit{non-sample-level} tokenizers aggregate information across time, compressing temporal and spectral structure into higher-level tokens (see Section II for a review of prior work). While both approaches have been adopted in recent studies (cf. \cite{Ansari_2024, Fang_2024}), most tokenization strategies were originally developed for non-biological time series such as retail, finance, or epidemiology, whose statistical properties differ substantially from those of neural signals. This lack of modality-specific design principles raises the question of whether these tokenization strategies faithfully capture the structure of M/EEG data, which exhibit oscillatory dynamics, structured spectral organization, and approximately Gaussian amplitude distributions \cite{Quinn_2019}. Furthermore, current practice lacks consensus among the field, with tokenization choices often inherited from prior work or driven by architectural convenience rather than systematic evaluation.

To date, no study has systematically examined how tokenization strategies affect representational fidelity, generative behavior, and downstream task performance of the subsequent foundation model for neural time series. In this work, we address this gap by evaluating tokenization methods along two complementary axes. First, we assess their ability to represent continuous neural signals in a low-dimensional discrete space without information loss, quantified via reconstruction accuracy. Second, we pretrain a generative pretrained transformer (GPT)-style foundation model \cite{Radford_2018, Brown_2020} and examine how tokenization influences the model behavior by evaluating the (i) token prediction accuracy, (ii) biological plausibility of the generated neural data, (iii) capacity to capture subject-specific signatures and inter-subject variability, and (iv) performance on downstream decoding tasks under zero-shot and fine-tuning settings.

We focus exclusively on sample-level tokenization and defer the analysis of non-sample-level approaches to future work. Although non-sample-level tokenizers are widely employed in M/EEG foundation modeling \cite{Jiang_2024, Ouahidi_2025}, sample-level tokenization offers several conceptual and practical advantages. First, by avoiding temporal compression, it preserves the temporal and spectral resolution of the signal. When applied independently to each sensor or channel, it also retains spatial resolution. This design delegates the modeling of spatiotemporal structure entirely to the transformer-based foundation model, maximizing the information available to the model. Moreover, sample-level tokenization renders tokens themselves directly interpretable by ensuring precise temporal alignment between tokens and the underlying signals, which is particularly important for M/EEG analyses. Second, sample-level tokenization is, in theory, expected to generalize more robustly across datasets, as it avoids encoding temporally dense or dataset-specific structure within individual tokens. This property may further reduce sensitivity to variations in preprocessing pipelines, acquisition hardware, or source reconstruction methods. Finally, to the best of our knowledge, all existing sample-level tokenizers for neural data are non-learnable and rely on fixed discretization schemes. We introduce the first learnable (i.e., data-adaptive), sample-level tokenizer designed specifically for M/EEG data and compare it against established non-learnable baselines.

The main contributions of this paper are threefold:
\begin{itemize}
    \item We propose a learnable, sample-level tokenization framework tailored to the statistical and spectral characteristics of continuous M/EEG signals.
    \item Using MEG data, we present the first comparative evaluation of sample-level tokenization strategies for transformer-based LNMs.
    \item Using a controlled experimental setting where the datasets and model architecture are held constant, we show that simple, fixed discretization schemes perform comparably to learnable tokenizers across most evaluation criteria, with the exception of subject fingerprinting, where learnable tokenizers yield consistent improvements.
\end{itemize}
 
The remainder of this paper is organized as follows. Section II reviews related work on tokenization strategies for time series data. Section III describes the tokenizers and foundation model used in this study. Section IV details the experimental setup and presents comparative results of tokenizer evaluation. Section V discusses the implications of these findings, and Section VI concludes with a summary of the main results.

\section{Background and Related Work}

In this paper, we categorize time-series tokenization methods based on the temporal granularity of the individual tokens. In what follows, we review non-sample-level and sample-level tokenization strategies proposed in the general time-series literature and discuss their relevance to neural signal modeling.

\subsection{Non-Sample-Level Tokenization}

The three most widely used non-sample-level tokenization strategies are patching, time-frequency transforms, and vector quantization.

\textit{Patching} divides a continuous time series into fixed-length, non-overlapping or partially overlapping temporal segments. Each patch is treated as a token and an input to a concomitant foundation model. This approach was introduced in time-series modeling \cite{Nie_2023, Das_2024, Woo_2024} and later adopted in M/EEG LNMs \cite{Wang_2025b, Ouahidi_2025, Cui_2024, Ansari_2025}. Given that time series often exhibit strong local temporal dependencies, patching the data can reduce sequence length and computational cost without losing much detail. However, as patching primarily captures short-range patterns, global dependencies and multi-scale dynamics are left to be inferred by downstream transformers. In addition, hard patch boundaries can introduce artifacts, and frequency information is only implicitly represented.

\textit{Time-frequency (TF) transforms} map raw signals into joint TF representations such as continuous wavelet transforms (CWT) or short-time Fourier transform (STFT) spectrograms, aiming to preserve original waveforms and information across frequency scales. These transforms are either operated on time series directly \cite{Masserano_2025} or may be further patched like images in a Vision Transformer (ViT)-style \cite{Luo_2025}. Here, tokens typically correspond to TF coefficients, optionally compressed or made sparse over the time and frequency domains. Many time-series data exhibit structure that is easier to model in TF space, where frequency decomposition exposes periodicities, transients, and multi-scale dynamics that raw patches dilute. As many signals have sparse frequency representations, it can also reduce token count without losing much detail. Moreover, frequency-domain features can improve cross-sensors or cross-domain generalization and remain interpretable due to their correspondence with known spectral patterns. The main drawbacks are the need for more intricate and domain-specific preprocessing, as representation quality is sensitive to the choice of relevant transform and windowing hyperparameters.

\textit{Vector quantization (VQ)}, typically implemented via Vector Quantized-Variational AutoEncoder (VQ-VAE) \cite{van_den_Oord_2017} or Residual Vector Quantization (RVQ) \cite{Lee_2022}, learns a quantized latent space (i.e., discrete codebook), in which continuous time-series segments are encoded as quantizable dictionary embeddings (or tokens). This approach has been introduced in models such as TOTEM \cite{Talukder_2024} and adopted in M/EEG LNMs \cite{Jiang_2024, Xiao_2025, Barmpas_2025}. VQ\footnote{Some hybrid approaches integrate TF transforms with VQ, in which the dictionary is learned after signals are time-frequency-transformed \cite{Lee_2023} or composed from both time and frequency representations simultaneously \cite{Pradeepkumar_2025}.} enables efficient compression and scalable pretraining on large unlabeled data by converting long signals into short discrete sequences, as the learned codebook capture prototypical patterns that reflect complex local dynamics. However, codebook collapse or quantization error can degrade performance in tasks requiring high temporal precision (e.g., forecasting or anomaly detection).

Although the non-sample-level tokenization methods\footnote{Beyond these three families, motif-based compression methods such as Byte Pair Encoding (BPE) \cite{Götz_2025} are also noteworthy. In these approaches, signals are first scaled and quantized under a fixed discretization scheme, after which repeating motifs are extracted as tokens. Such methods retain fine-grained temporal resolution and computational efficiency comparable to sample-level tokenization, while additionally reducing redundant temporal repetitions.} differ greatly in their inductive biases, compression targets, and downstream objectives, they all share a common goal of summarizing continuous time series into informative, structured, and computationally efficient units that downstream models can readily exploit.

\subsection{Sample-Level Tokenization}

In time series modeling, sample-level tokenization offers several methodological benefits for the transformer-based foundation models over the non-sample-level approaches. Notably, discretizing continuous time series data enables direct adoption of modeling and training paradigms developed for Large Language Models (LLMs), without requiring modifications to the underlying architecture or learning objectives. Tokenization allows us to employ the cross-entropy as our loss function for the foundation model, which is known to provide superior numerical stability and more robust convergence behavior during training than regression-based objectives such as mean squared error (MSE) \cite{Géron_2022, Golik_2013}. Furthermore, having the foundation model output as a discrete probability distribution circumvents restrictive assumptions regarding the shape of the distribution (i.e., we avoid parametric distributions) and allows for the expression of arbitrary, potentially multimodal data distributions \cite{Ansari_2024, Gruver_2023}.

To minimize computational overhead and retain architectural simplicity, existing sample-level tokenizers typically employ fixed, non-learnable quantization schemes with no trainable parameters. A representative example is Chronos \cite{Ansari_2024}, which applies mean scaling followed by uniform binning to map continuous observations onto a finite vocabulary of discrete tokens. Transformer-based LLMs trained directly on such tokenized sequences have been shown to achieve competitive zero-shot performance on previously unseen datasets compared to task-specific time series forecasting models. By modeling each time point as a categorical variable over discrete tokens, this approach effectively reduces time series regression to a classification problem while avoiding the need for time series-specific architectural components or features.

It is important to note, however, that, while mean scaling with uniform binning is effective for many general-purpose time series domains, it is suboptimal for neural data due to their unique statistical characteristics and oscillatory dynamics. Following \cite{Ansari_2024}, in the case of neural time series z-score normalization with quantile-based binning provides a more appropriate tokenization strategy. Unlike mean scaling, standardization avoids placing disproportionate semantic emphasis on zero values, which is preferable for neural data where zeroes do not necessarily represent a meaningful baseline. Quantile binning ensures an approximately uniform distribution of token counts, allowing the model to learn from balanced token occurrences during pretraining.

A closely related quantization strategy is the $\mu$-law companding transform \cite{ITUT_1988}, originally developed for audio signal processing and later adopted in WaveNet \cite{van_den_Oord_2016} for audio generation. This method applies max-absolute scaling followed by logarithmic $\mu$-law compression and fixed-level quantization. Because the $\mu$-law companding allocates higher resolution to small-magnitude values while more aggressively compressing larger ones, the resulting bins are non-uniform in the original data space, concentrating around the center of the distribution to preserve subtle, high-density structure in the signal. However, as the companding algorithm is designed for audio signals and assumes fixed, symmetric minimum and maximum values, its application to neural time series requires explicit signal clipping to satisfy these assumptions \cite{Csaky_2024}.

\section{Methods}

\subsection{Overview}

\begin{figure}[!t]
\centerline{\includegraphics[width=\columnwidth]{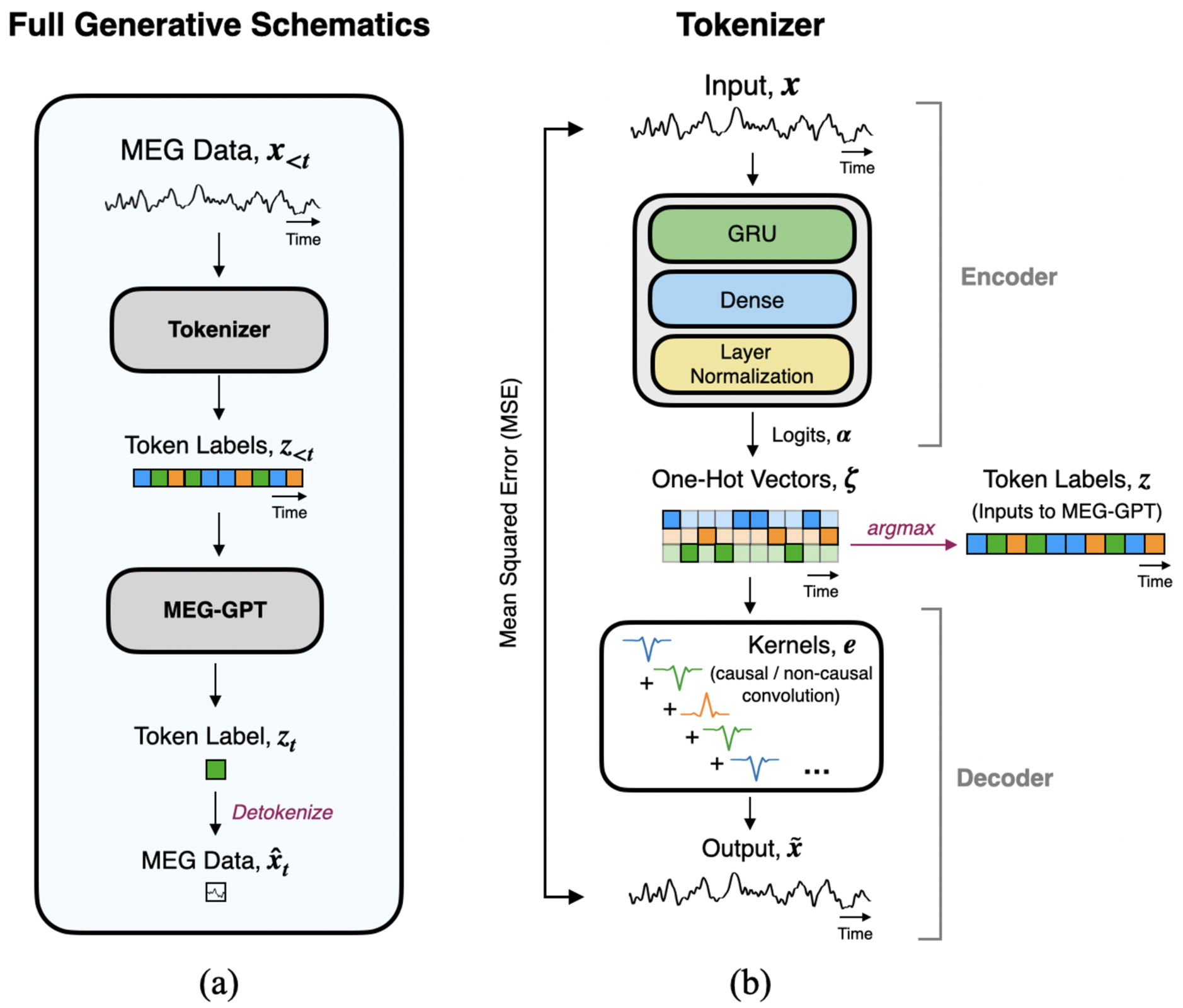}}
\caption{Overview of the foundation modeling framework and tokenizer architecture. (a) Schematic of the full generative training pipeline for the MEG-GPT foundation model. (b) Architecture of the learnable MEG tokenizer.}
\label{fig1}
\end{figure}

In this paper, we evaluate different sample-level tokenization strategies by comparing their ability to compactly represent continuous data as discrete tokens and the downstream performance of foundation models trained with these tokenizers. The complete foundation modeling pipeline is shown in \hyperref[fig1]{Fig.~1(a)}. MEG signals are first converted into sequences of discrete token labels, which are then provided as input to the foundation model. In this study, we employ MEG-GPT \cite{Huang_2025} as this foundation model (see Section~III-D). As a GPT model, MEG-GPT is trained using a self-supervised next-token prediction objective \cite{Radford_2018, Bommasani_2022}. The following sections describe the tokenizers and MEG-GPT model used in this work.

\subsection{Learnable MEG Tokenizers}

To accommodate the characteristics of MEG data, we introduce a learnable sample-level tokenizer specifically designed to capture the statistical and spectral structure of continuous MEG signals (\hyperref[fig1]{Fig. 1(b)}).

The proposed tokenizer maps each continuous MEG time series into a corresponding sequence of discrete tokens, with tokenization performed independently for each channel or brain region. It learns a finite set of short, recurring temporal patterns (or tokens) by training under an autoencoder-based framework. Our tokenization strategy is designed to preserve the full temporal and spatial resolution of the MEG data, leaving the modeling of spatiotemporal dependencies entirely to MEG-GPT. To this end, the tokenizer is trained without additional regularization, yielding a near-lossless transformation of the input signals. This design choice facilitates interpretability of the learned tokens and preserves exact temporal alignment between the raw and tokenized time series, which is critical for post hoc MEG analyses.

Conceptually, the tokenizer can be viewed as a VQ-VAE-like network \cite{van_den_Oord_2017} in which the codebook is embedded within the decoder rather than implemented as an explicit quantization module. Furthermore, as a standard autoencoder, it does not require stochastic sampling or a straight-through gradient estimator during training.

\subsubsection{Encoder}

In our tokenizer, the encoder maps continuous, real-valued MEG time series $x$ into a sequence of categorical token labels $z$, where each label is a unique integer index. Specifically, for a single-channel MEG signal $x \in \mathbb{R}^T$ with $T$ time samples, the encoder outputs a sequence of one-hot vectors $\zeta \in \mathbb{R}^{V \times T}$, where each vector indicates the inferred token at a given time step. Here, $V$ denotes the size of token vocabulary, which is predefined as a hyperparameter.

To infer $\zeta$, the encoder predicts a sequence of logits $\alpha \in \mathbb{R}^{V \times T}$ as
\begin{equation}
    \alpha = \mathrm{Encoder}(x).
\end{equation}
Applying a softmax over the $V$-dimension of $\alpha$ yields token probabilities $p \in [0, 1]^{V \times T}$. The encoder architecture comprises a single Gated Recurrent Unit (GRU) layer \cite{Cho_2014}, followed by a fully connected linear dense layer and layer normalization \cite{Ba_2016}. The inferred token at each time step $t$ is then formulated as a one-hot vector such that
\begin{equation}
    \zeta_t = \mathrm{OneHot}(\arg\max(\alpha_t)), \quad \forall t \in \{1, \dots, T\}.
\end{equation}
Equivalently, the sequence of inferred token labels can be expressed in discrete form as $z = \arg\max(\zeta)$, where $z \in \mathbb{R}^{T}$.

\subsubsection{Decoder}

After encoding MEG data into token labels, the decoder reconstructs the continuous data from the inferred one-hot vectors $\zeta$. To model temporal smoothness of the time-series, we employ a set of one-dimensional (1D) convolution kernels, which can be interpreted as token-specific filters (or a learned \textit{dictionary}). In this context, the one-hot vectors act as impulses that are convolved with the token kernels and summed to produce the reconstructed signal.

Let $e \in \mathbb{R}^{d_{\mathrm{token}} \times V}$ be the set of learnable token kernels, with each kernel a vector of dimension $d_{\mathrm{token}}$. The reconstructed signal $\tilde{x}_t$ at time $t$ is computed as a weighted sum of all token kernels:
\begin{equation}
    \begin{split}
        \tilde{x}_t &= \mathrm{Decoder}(\zeta) \\
        &= \sum^V_{v=1} \left(w_v \left(\sum^{\left\lceil \frac{d_{\mathrm{token}}-1}{2} \right\rceil}_{\tau=-\left\lfloor \frac{d_{\mathrm{token}}-1}{2} \right\rfloor} e_{v, \tau}\zeta_{v,t-\tau} \right) + b_v\right)    
    \end{split}
\end{equation}
where $w_v$ and $b_v$ are learnable weights and biases, respectively, used to combine the outputs of 1D convolutions.

Note that the formulation above corresponds to a \textbf{noncausal} convolution, as the reconstruction at time $t$ depends on both past and future token assignments. To make the tokenizer fully \textbf{causal}, the decoder can alternatively apply each kernel using only the current and preceding tokens:
\begin{equation}
    \tilde{x}_t = \sum^V_{v=1} \left(w_v \left(\sum^{d_{\mathrm{token}}-1}_{\tau=0} e_{v, \tau}\zeta_{v,t-\tau} \right) + b_v\right).
\end{equation}
This causal formulation prevents the decoder from accessing future information, making the tokenizer putatively suitable for applications that require real-time or temporally constrained processing.

\subsubsection{Training the tokenizer}

The tokenizer is trained using stochastic gradient descent with the Adam optimizer \cite{Kingma_2015} by minimizing the MSE between the input signal $x$ and its reconstruction $\tilde{x}$. A key consideration in this training procedure is that the computation of the one-hot vectors $\zeta$ relies on an $\arg\max$ operation, which enforces categorical token labels but is non-differentiable. As a result, gradients cannot be backpropagated through this operation, hindering optimization of the encoder parameters.

To address this issue, we adopt an annealed relaxation of the hard token assignment and define a one-hot vector $\zeta$ as:
\begin{equation}
    \zeta_t = (1 - \kappa) \cdot \mathrm{OneHot}(\arg\max(\alpha_t)) + \kappa \cdot \mathrm{Softmax}(\alpha_t)
\end{equation}
where $\kappa \in [0, 1]$ is an annealing coefficient. During training, $\kappa$ is initialized to $1$ and linearly decayed to $0$, allowing the model to transition smoothly from a soft, differentiable representation to a hard categorical representation. At inference time, $\kappa$ is fixed to $0$ to produce discrete token assignments.

\subsubsection{Token refactorization}

In practice, not all tokens in the predefined vocabulary are necessarily utilized during reconstruction. In this case, we perform a post-training token refactorization step \cite{Huang_2025} in which unused tokens are removed from the vocabulary and relabeled in descending order prior to training the foundation model. After refactorization, the effective vocabulary consists of $V^* + 1$ tokens, where $V^{*}$ denotes the number of retained tokens and the additional token\footnote{This is conceptually similar to an out-of-vocabulary (OOV) or unknown (UNK) token in the context of NLP.} is reserved for mapping out-of-distribution samples that fall outside the bin ranges observed during tokenizer training.

\subsection{Baseline Tokenizers}

To benchmark the proposed tokenizer, we implemented two non-learnable, sample-level baseline methods surveyed in Section~II-B. Both baselines follow a fixed quantization pipeline in which the input time series is first scaled using an affine transformation, $x'_t = (x_t - m) / s$, and subsequently discretized according to specified quantization rules.

\subsubsection{\texorpdfstring{$\mu$-Transform Tokenizer}{μ-Transform Tokenizer}}

The first baseline is the $\mu$-transform tokenizer \cite{van_den_Oord_2016}. To prevent extreme values from disproportionately influencing the bin boundaries, the input signals are clipped prior to scaling. The time series is then normalized using max-absolute scaling by setting $m = 0$ and $s = \max_{1 \le t \le T} \lvert x_t \rvert
$, yielding $x' \in [-1, 1]$. The normalized signal is subsequently compressed using the $\mu$-law companding function
\begin{equation}
    F(x) = \mathrm{sgn}(x)\frac{\ln(1 + \mu |x|)}{\ln(1 + \mu)}, \quad -1 \leq x \leq 1,
\end{equation}
after which the compressed values are discretized into bins. Owing to the nonlinear nature of $\mu$-law companding, larger-magnitude values are compressed more aggressively, resulting in non-uniform bin widths in the original signal space. Bins are narrower near the mean—where neural time-series samples are typically concentrated—and wider in the tails, which often contain outliers.

\subsubsection{Standard-Quantile Tokenizer}

The second baseline is the standard-quantile (SQ) tokenizer \cite{Ansari_2024}. Just like the $\mu$-transform tokenizer, we start by clipping the input signals. Then, each time series is standardized via $z$-score normalization, with
\begin{equation}
    m = \frac{1}{T} \sum_{t=1}^{T} x_t,
    \quad
    s = \sqrt{\frac{1}{T} \sum_{t=1}^{T} (x_t - m)^2}.
\end{equation}
Next, the standardized values are assigned to bins defined by empirical quantiles, such that each bin contains approximately the same number of samples. This procedure yields a near-uniform token frequency distribution and promotes evenly balanced exposure of the foundation model to different value ranges.

\subsubsection{Detokenization and Out-of-Distribution Handling}

Both baseline tokenizers handle out-of-distribution samples by assigning values outside the predefined range to the outermost bins, corresponding to the lowest or highest token indices depending on the sign. Detokenization is performed by mapping each token back to the center of its associated bin, followed by inversion of the corresponding normalization transform. Specifically, for the $\mu$-law tokenizer, the inverse $\mu$-law transform and inverse max-absolute scaling are applied, whereas for the SQ tokenizer, the inverse $z$-score normalization is used.

The scaling parameters $m$ and $s$, as well as the bin boundaries, are estimated from the tokenizer training dataset and held fixed during inference. These parameters are subsequently used to tokenize both the test data and the data employed for the foundation model training.

\subsection{Foundation Model Network Architecture}

\begin{figure}[!t]
\centerline{\includegraphics[width=\columnwidth]{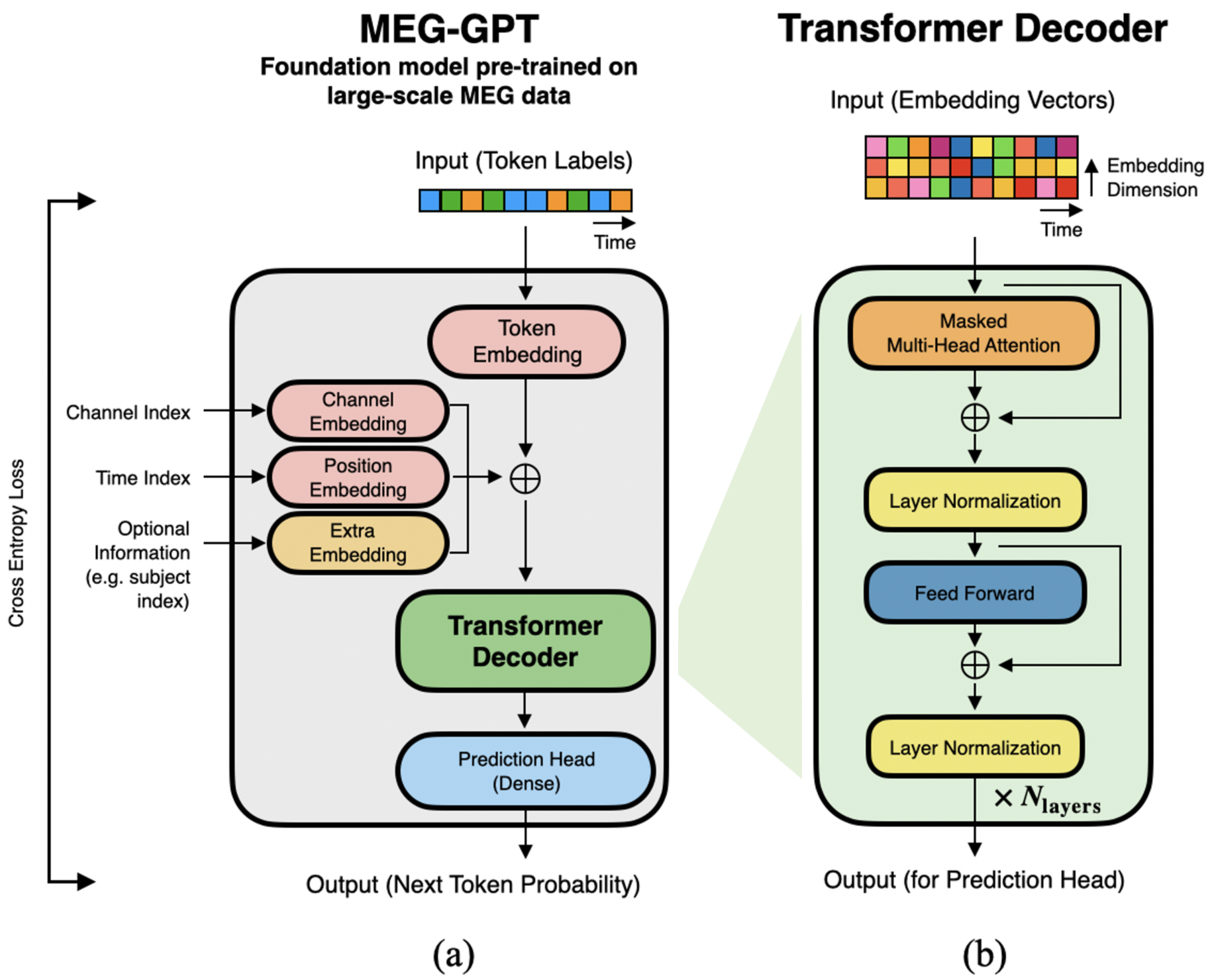}}
\caption{MEG-GPT foundation model architecture. (a) High-level overview of the model architecture. (b) Detailed structure of the transformer decoder component.}
\label{fig2}
\end{figure}

As mentioned above, we adopt MEG-GPT as the foundation model for evaluating different tokenization strategies. MEG-GPT was selected because, to our knowledge, it is the only LNM explicitly designed and trained with sample-level tokenization for neural time-series data. Although prior work such as \cite{Ansari_2024} applies sample-level tokenizers to existing LLM architectures, those models were not pretrained on neural signals such as M/EEG. Moreover, adapting sample-level tokenization to models originally designed using non-sample-level approaches typically requires modifications to the learning objective or network architecture, and the impact of such changes on model behavior is unclear. On this account, MEG-GPT can serve as a principled and controlled benchmark for comparing sample-level tokenizers in transformer-based LNMs for neural time series.

MEG-GPT is a foundation model developed to learn generalizable, abstract representations from large-scale MEG data. The overall architecture of MEG-GPT is illustrated in \hyperref[fig2]{Fig.~2(a)}, with its core component being a transformer decoder \cite{Vaswani_2017} (\hyperref[fig2]{Fig.~2(b)}) that parametrizes temporal dependencies among tokens via self-attention. This mechanism allows the model to attend to different parts of the past in a context-dependent manner.

From a modeling perspective, MEG-GPT can be viewed as a nonlinear autoregressive (AR) model implemented with a transformer architecture. Unlike classical AR models, however, the AR weights can change as a function of the data through the multi-head attention mechanism. The model predicts each subsequent token from preceding tokens by minimizing the cross-entropy loss between the predicted token distribution and the ground-truth labels using teacher forcing. In brief, input token labels are first represented as learned embedding vectors for each time point and channel, encoding token characteristics and grouping similar tokens together in the embedding space. The transformer decoder then learns statistical dependencies between these embeddings to capture temporal information from neural dynamics. Finally, the decoder outputs are projected through a linear layer to produce logits that define the next-token probability distribution. For detailed architectural and implementation details, please refer to \cite{Huang_2025}.

\subsection{Synthetic Data Generation}

One approach we use for assessing the effect of tokenization on the foundation model is to examine whether MEG data generated by MEG-GPT resemble real signals and preserve their key biological properties. Using MEG-GPT as a generative model, we synthesized new MEG data \cite{Huang_2025}. First, an initial prompt sequence of tokens was constructed by sampling from a categorical distribution weighted by token occurrences in the training set. This prompt, together with auxiliary inputs such as subject indices and channel labels, was provided to MEG-GPT. The model then generated subsequent tokens autoregressively using nucleus sampling with top-$p=0.99$ \cite{Holtzman_2020}. Finally, the generated token sequences were passed through the decoder of the pretrained tokenizer to convert discrete token labels back to continuous synthetic MEG time series.

\section{Experiments and Results}

In this section, we present a comparative evaluation of sample-level tokenizers across five complementary criteria. We first describe the datasets and model training procedures, including the training configurations and pretraining setup of the MEG-GPT foundation model. We then report qualitative and quantitative analyses of (i) reconstruction fidelity of continuous neural signals from discrete representations, (ii) token prediction performance, (iii) the generative quality of synthetic MEG signals, (iv) the ability to capture subject-specific information, and (v) performance on downstream neural decoding tasks.

\subsection{Datasets}

We used three publicly available MEG datasets to train and evaluate the tokenizers and MEG-GPT models:

\subsubsection{Dataset I}

The Cambridge Centre for Ageing and Neuroscience (Cam-CAN) dataset \cite{Shafto_2014} consists of resting-state, eyes-closed MEG data collected using a 306-channel Elekta scanner at a sampling frequency of 1 kHz. It contains scans from 612 healthy participants (aged 18-88 years; 310 males, 302 females), each lasting approximately 8.5 minutes. 50 subjects were randomly subsampled from the entire dataset to train tokenizers, and the remaining subjects were used to evaluate tokenizer generalization to unseen participants. This dataset was also used for pretraining and evaluating MEG-GPT models.

\subsubsection{Dataset II}

The Nottingham MEGUK dataset\footnote{https://meguk.ac.uk/database/} consists of resting-state, eyes-open MEG data collected using a 275-channel CTF scanner at a sampling frequency of 1.2 kHz. It contains scans from 65 healthy subjects (aged 18 to 60+ years; 31 males, 34 females), each lasting approximately 5 minutes. This dataset was exclusively used to assess the zero-shot generalization capability of the tokenizers.

\subsubsection{Dataset III}

The Wakeman-Henson dataset \cite{Wakeman_2015} is a task-based MEG dataset acquired using a 306-channel Elekta scanner at a sampling frequency of 1 kHz. It contains recordings from 19 healthy participants (aged 23-37 years; 11 males, 8 females), each scanned across six sessions. During each session, participants were presented with visual stimuli belonging to three categories: famous, unfamiliar, and scrambled faces. Each recording session lasted approximately 7.5 minutes and contained about 200 trials, with stimulus categories evenly distributed. To ensure task engagement, participants were instructed to indicate via button press whether each stimulus appeared symmetric. This dataset was used for tokenizer evaluation under task-evoked conditions and for downstream decoding experiments.

While recent M/EEG LNMs have focused primarily on improving downstream classification accuracy, our evaluation extends beyond predictive performance to assess the biological plausibility of the generated signals, including their spatiotemporal organization and spectral characteristics. This necessitates the use of datasets that have been widely adopted in neuroscience research, thereby providing a well-established foundation for biological and neuroscientific interpretation. The three datasets above satisfy these requirements and collectively provide diverse recording conditions, tasks, and acquisition systems for comprehensive model evaluation.

\begin{table*}[!t]
\centering
\caption{Model Performance Metrics of Tokenizers and MEG-GPT}
\label{tab1}
\begin{tabular}{ll|ccccccc}
    \toprule
    & \multicolumn{1}{c}{} & \textbf{Causal} & \textbf{Noncausal} & \multicolumn{4}{c}{\textbf{\bm{$\mu$}-Transform}} & \textbf{Standard Quantile} \\
    \midrule
    \multirow{4}{*}{\textbf{Tokenizer}} & \# Runs & 10 & 10 & \multicolumn{4}{c}{-} & - \\
    & Mean $\pm$ Std. & \textbf{0.0258 $\pm$ 8.58e-4} & 0.0277 $\pm$ 5.33e-3 & \multicolumn{4}{c}{-} & - \\
    & Best Final Loss & 0.0245 & \textbf{0.0170} & \multicolumn{4}{c}{-} & - \\
    \cmidrule{2-9}
    & \# Tokens & 97 & 121 & 256 & 182 & 108 & 54 & 108 \\
    \midrule
    \multirow{4}{*}{\textbf{MEG-GPT}} & Train Loss & 1.7540 & 1.8740 & 3.6158 & 3.3824 & 3.0263 & 2.5492 & 3.2217 \\
    & Train Top 1 Acc. & 0.3353 & 0.3426 & 0.0721 & 0.0873 & 0.1172 & 0.1728 & 0.0944 \\
    & Validation Loss & 1.7556 & 1.8760 & 3.6165 & 3.3829 & 3.0276 & 2.5501 & 3.2224 \\
    & Validation Top 1 Acc. & 0.3350 & 0.3421 & 0.0722 & 0.0874 & 0.1173 & 0.1728 & 0.0945 \\
    \bottomrule
    \multicolumn{9}{l}{\footnotesize Boldface indicates the best value between the learnable tokenizers. Std.: standard deviation, Acc.: accuracy.}
\end{tabular}
\end{table*}

\subsection{Data Preparation}

All datasets were preprocessed and source-reconstructed using the \texttt{osl-ephys} toolbox \cite{van_Es_2025}. Detailed dataset-specific preprocessing pipelines are reported in \cite{Huang_2025} for Datasets I and III and in \cite{Cho_2025} for Dataset II; for completeness, we summarize the steps here.

\subsubsection{Preprocessing}

Raw MEG recordings were band-pass filtered between 0.5 and 125 Hz and denoised to remove power-line interference at 50 and 100 Hz. The data were subsequently downsampled to 250 Hz. Noisy time segments and channels were automatically detected and excluded using the generalized extreme Studentized deviate (GESD) algorithm \cite{Rosner_1983}. Independent component analysis (ICA) was then performed using FastICA \cite{Hyvarinen_1999} to decompose the sensor-level signals into 64 components. Components exhibiting high correlation with electrooculogram (EOG) or electrocardiogram (ECG) signals were identified as artifacts and removed.

\subsubsection{Source Reconstruction}

The preprocessed sensor-level data were co-registered and source-reconstructed onto an 8-mm isotropic grid using a volumetric linearly constrained minimum variance (LCMV) beamformer \cite{Veen_1988}. The resulting voxel-wise time series were parcellated into 52 anatomically defined regions \cite{Kohl_2023}. To mitigate spurious inter-parcel correlations and reduce source leakage, we applied the symmetric multivariate leakage reduction technique \cite{Colclough_2015}, which removed all zero-lag correlation between parcel time courses.

Following source reconstruction, each parcel time series was temporally standardized by subtracting its mean and dividing by its standard deviation along the time dimension. All subsequent analyses were conducted on these standardized source-level signals.

\subsection{Model Training}

Our models and methods are implemented in Python 3.10 using the TensorFlow library (version 2.11) with two NVIDIA A100 or V100 GPUs. Training took  about 0.66 and 150 GPU hours for the learnable tokenizer and MEG-GPT, respectively. The performance metrics for the learnable tokenizers and MEG-GPT models are summarized in \hyperref[tab1]{Table I}.

\subsubsection{Tokenizers}

Both causal and noncausal learnable tokenizers were trained using the Adam optimizer with a learning rate of 1e-5 and momentum parameters $\beta_1 = 0.9$ and $\beta_2 = 0.999$. Training was performed with a batch size of 32 and a sequence length of 200. The encoder consisted of a GRU layer with 128 hidden units, while the decoder employed 1D convolution kernels of width 10. All models were trained for 40 epochs. The annealing coefficient $\kappa$ was linearly annealed over the same number of epochs.

To account for stochasticity in training, each tokenizer variant was trained independently 10 times, and the model achieving the lowest final loss was selected for subsequent analyses. The choice of 40 training epochs was determined via grid-search hyperparameter tuning over $\{10, 40, 60\}$ epochs. The initial token vocabulary size was set to $V = 128$; however, due to token refactorization during training, the optimal runs resulted in effective vocabulary sizes $V^{\ast} = 96$ and $V^{\ast} = 120$ for the causal and noncausal tokenizers, respectively.

For the non-learnable $\mu$-transform and SQ tokenizers, a fixed vocabulary size of 108 tokens—corresponding to the average $V^{\ast}$ across the learnable tokenizers—was used. Unlike learnable tokenizers, non-learnable approaches rely on center values of bin edges during detokenization, which can introduce discontinuities, particularly when the vocabulary size is small. Moreover, the absence of a learning mechanism requires the vocabulary size to be manually specified based on prior knowledge of the data modality and experimental paradigm. Hence, to further examine the impact of vocabulary size, the $\mu$-transform tokenizer was additionally evaluated with 256, 182, and 54 tokens. As these tokenizers contain no trainable parameters, they were applied to the data only once.

Tokenizer training was conducted using parcel time courses from the first 50 subjects of the Cam-CAN dataset. The testing set included the remaining Cam-CAN subjects as well as the full Wakeman-Henson and Nottingham MEGUK datasets.

\subsubsection{MEG-GPT}

MEG-GPT models were trained separately using each tokenizer variant. For each subject in the Cam-CAN dataset, the tokenized data were split into training and validation sets using a nine-to-one split. To ensure fair comparison across tokenizer conditions and eliminate variability due to stochastic initialization, a fixed random seed was used for model pretraining, synthetic data generation, and all post-hoc analyses. Each MEG-GPT configuration was trained only once per tokenizer.

All MEG-GPT models were trained with the Adam optimizer using a learning rate of 1e-5 and momentum parameters $\beta_1 = 0.9$ and $\beta_2 = 0.999$. A batch size of 8 and a receptive field (sequence length) of 80 tokens were used, with the dimension of input embedding vectors as 400. The model consisted of four transformer decoder layers, each with four self-attention heads. The prediction head comprised a feedforward network with 400 hidden units and a dropout rate of 0.2, followed by a leaky ReLU activation and a dropout rate of 0.2. A complete list of hyperparameters is provided in the accompanying GitHub repository.

\begin{figure}[!t]
\centerline{\includegraphics[width=\columnwidth]{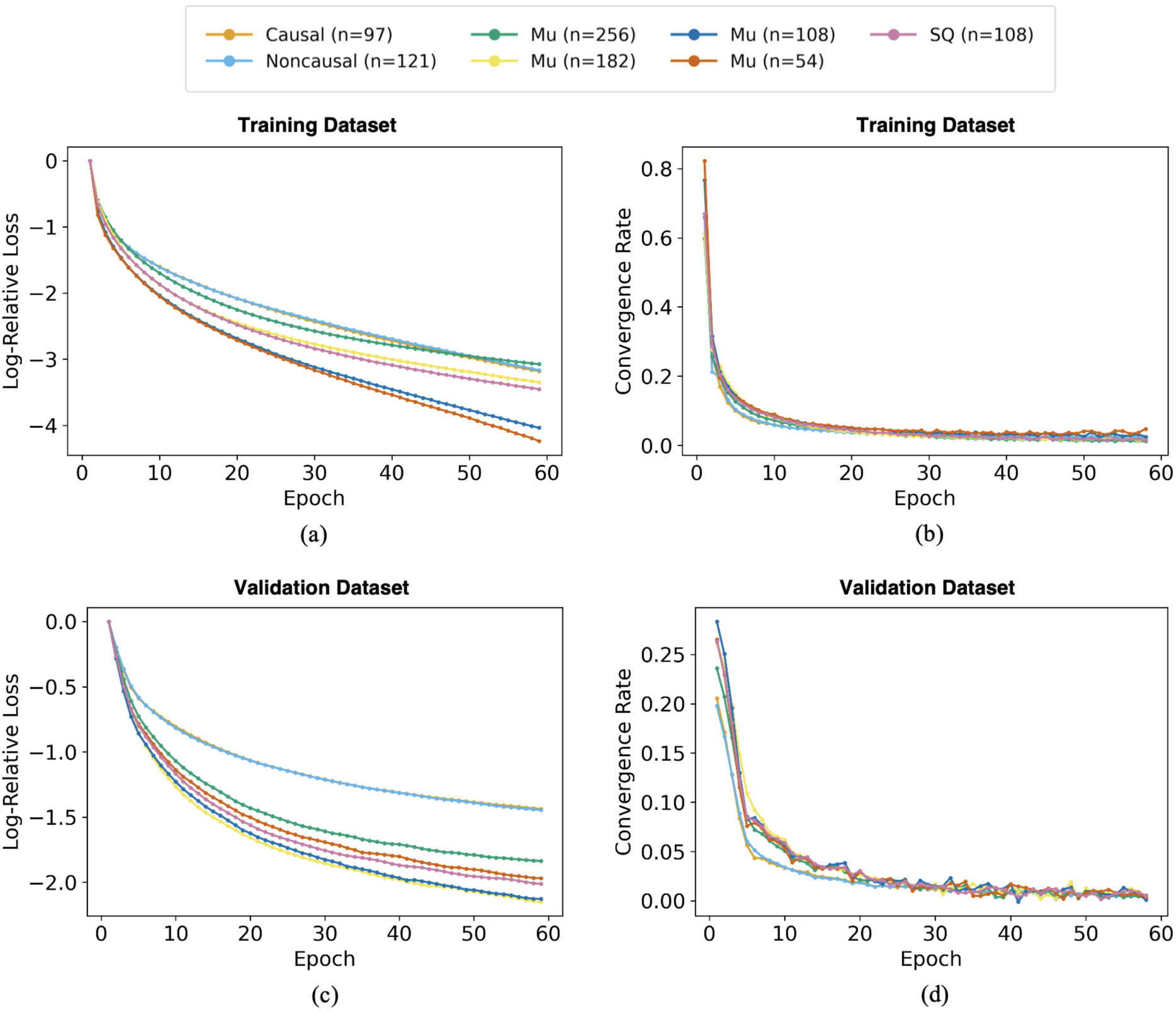}}
\caption{Convergence behavior of MEG-GPT models across tokenizer types on training and validation datasets. $n$ denotes the vocabulary size (i.e., number of unique tokens) for each tokenizer.}
\label{fig3}
\end{figure}

Because applying different tokenizers results in distinct sets of token vocabulary, the training loss and accuracy values across MEG-GPT models are not directly comparable. For this reason, performance metrics in \hyperref[tab1]{Table I} are reported without highlighting best-performing values. Instead, model comparability was assessed by examining loss convergence behavior. Training and validation loss curves were first smoothed using a Savitzky-Golay filter \cite{Savitzky_1964}, and their asymptotic values were estimated via nonlinear least-squares fitting using the Levenberg-Marquardt algorithm \cite{Levenberg_1944, Marquardt_1963}. We then computed the log-relative loss
\begin{equation}
    \log(\tilde{L}_t) = \log \left(\frac{L_t - L_\infty}{L_0 - L_\infty}\right),
\end{equation}
where $L_t$ denotes the loss at epoch $t$, $L_0$ the initial loss, and $L_\infty$ the estimated asymptotic loss. The instantaneous convergence rate was further defined as
\begin{equation}
    r_t = - \frac{d}{dt}\log(\tilde{L}_t).
\end{equation}
Across all tokenizer conditions, both training and validation losses decayed monotonically following exponential or polynomial trends (\hyperref[fig3]{Fig.~3(a), (c)}), with convergence rates approaching similar values by the end of training (\hyperref[fig3]{Fig.~3(b), (d)}). These results indicate that all MEG-GPT models converged comparably, thereby justifying their comparison in subsequent analyses.

\subsubsection{Zero-Shot Learning and Fine-Tuning}

For downstream task decoding, we applied the pretrained MEG-GPT models to the Wakeman-Henson dataset. Source-level parcel time courses were time-locked to stimulus onset and epoched. Task decoding was formulated as a four-class classification problem (famous, unfamiliar, scrambled faces, and button-press response) using a logistic regression classifier.

In the zero-shot setting, the MEG-GPT prediction head was discarded, and outputs with dimensions $(L, C, D)$ were extracted from the transformer decoder, where $L$ denotes sequence length, $C$ the number of parcels, and $D$ the latent dimension. Temporal information was collapsed via a linear projection over the sequence dimension. The resulting features were flattened across channel and latent dimensions, normalized using layer normalization, and passed through a linear layer to predict class probabilities.

For end-to-end fine-tuning, the same classification pipeline used in the zero-shot setting was integrated into the model, replacing the original prediction head with the logistic regression classifier. Unlike zero-shot decoding, the transformer decoder layers were fine-tuned jointly with the classifier. Token, channel, and position embedding layers were frozen. Subject embeddings were initialized by separately fine-tuning the pretrained MEG-GPT with all parameters frozen except the subject embedding layer.

A baseline decoding model was also evaluated using epoched parcel time courses with dimensions $(L, C)$. These features were flattened across time and channels, layer normalized, and classified using an identical logistic regression architecture.

In all three cases, classifiers were trained by minimizing the cross-entropy loss between predicted and ground-truth class labels. Performance was evaluated under two conditions: \emph{within-subject}, where the first five sessions of each participant were used for training and the sixth session for testing, and \emph{new-subject}, where all sessions from the first 18 participants were used for training and all sessions from the last participant were held out for testing.

\subsection{Signal Reconstruction Fidelity}

\begin{figure}[!t]
\centerline{\includegraphics[width=\columnwidth]{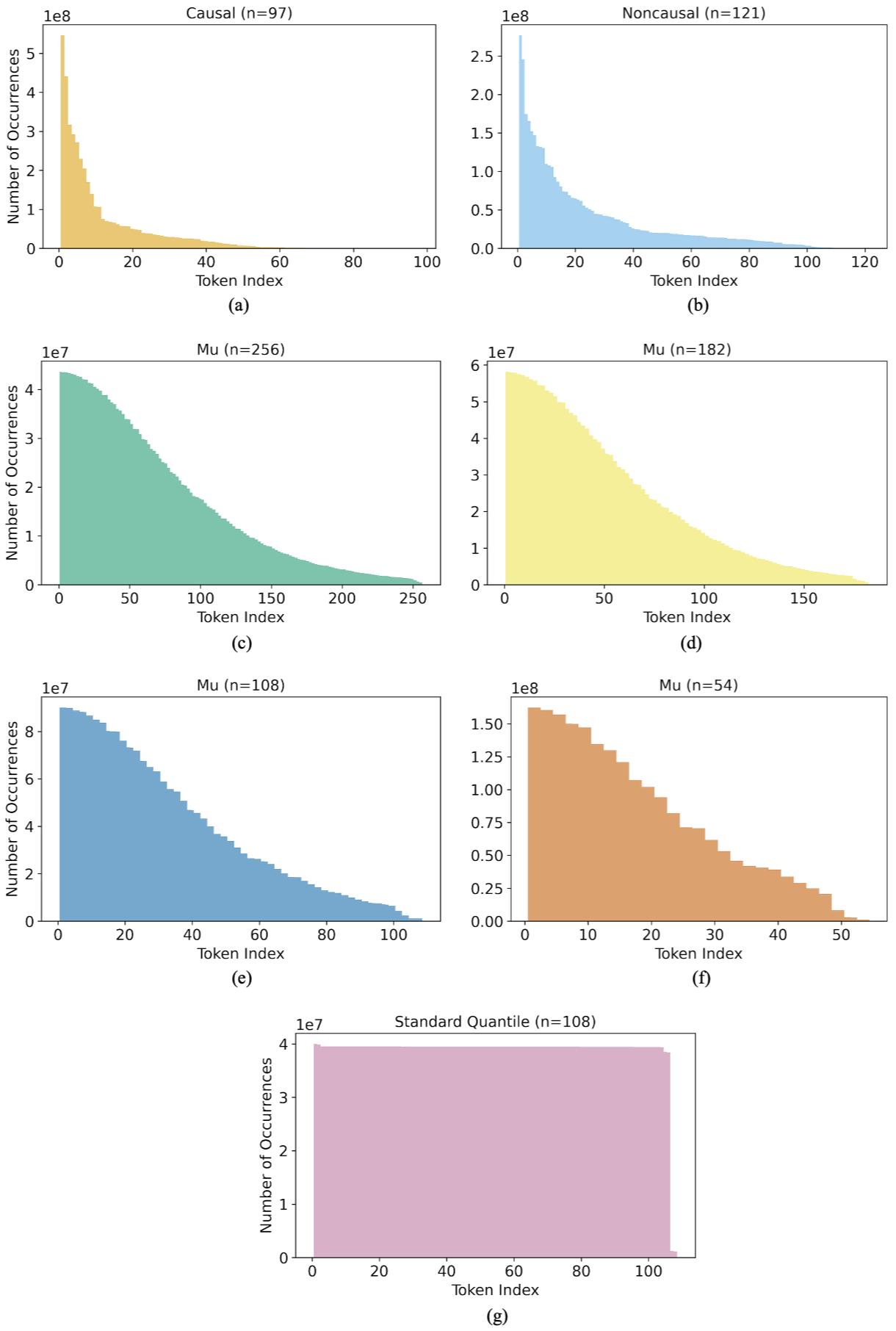}}
\caption{Distributions of input token counts for different tokenization methods. Token indices are sorted in descending order of frequency, and the value $n$ denotes the vocabulary size for each tokenizer.}
\label{fig4}
\end{figure}

The token distributions for the full Cam-CAN dataset are shown in \hyperref[fig4]{Fig.~4}. Both causal and noncausal tokenizers produced token distributions with steep exponential decay, indicating compact representations dominated by a small subset of frequently used tokens. The $\mu$-transform tokenizer exhibited a similar but more gradual decay, whereas the SQ tokenizer, by design, yielded an approximately uniform distribution. Notably, the learnable tokenizers converged to a parsimonious representation using approximately 100 tokens (97 for causal and 121 for noncausal).

\begin{figure}[!t]
\centerline{\includegraphics[width=\columnwidth]{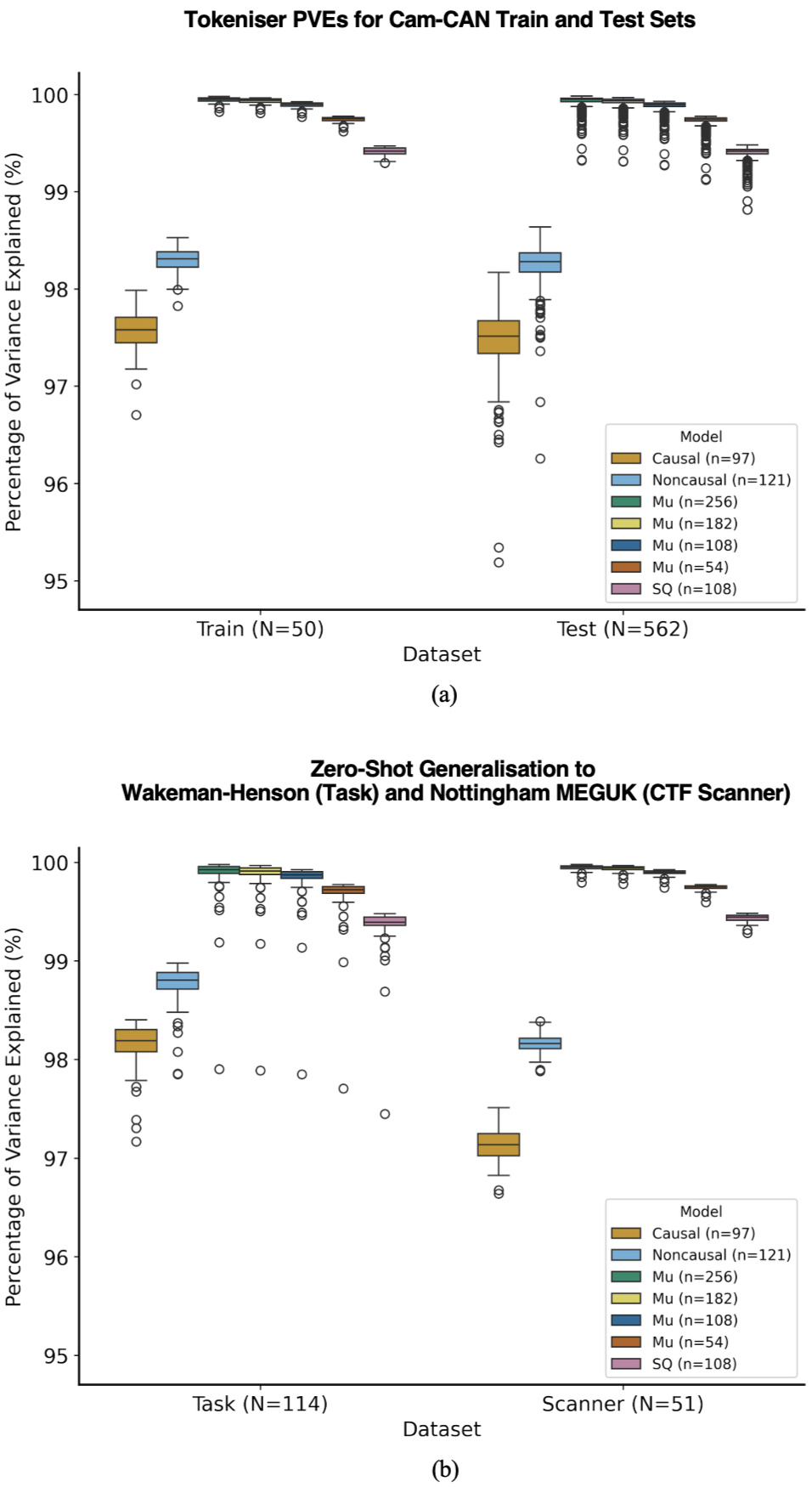}}
\caption{Reconstruction accuracy and zero-shot generalization performance of different tokenizers. (a) Subject-level PVE between original MEG recordings and data reconstructed from tokens for the Cam-CAN training and test datasets. (b) Same analysis as in (a) evaluated on previously unseen datasets acquired under a task paradigm or using a different MEG scanner. Here, $N$ denotes the number of subjects, and $n$ the number of unique tokens used by each tokenizer.}
\label{fig5}
\end{figure}

To quantitatively assess how well each tokenizer represents and reconstructs the original MEG data, we evaluated reconstruction fidelity using the subject-level percentage of variance explained (PVE) between the original and reconstructed (i.e., de-tokenized) parcel time courses, aggregated over both the time and channel dimensions. Across the Cam-CAN dataset, all tokenizers achieved high reconstruction accuracy, exceeding 97\% PVE on both training and test sets (\hyperref[fig5]{Fig.~5(a)}), with non-learnable baselines attaining slightly higher values ($>$99\% PVE). This outcome is in part expected, as non-learnable tokenizers effectively perform fixed binning akin to downsampling time series.

All tokenizers further demonstrated strong zero-shot generalization to unseen datasets. When applied to the Wakeman-Henson task data and the Nottingham MEGUK data acquired with a different MEG scanner, PVE values remained above 97\% and followed trends consistent with those observed in Cam-CAN (\hyperref[fig5]{Fig.~5(b)}).

In summary, our findings demonstrate that continuous MEG signals can be discretized into fewer than 130 tokens without substantial loss in reconstruction accuracy when the vocabulary size is learned adaptively. A similar result was observed for the $\mu$-transform tokenizer, where reducing the number of pre-specified tokens only modestly decreased PVE. Together, these results highlight that sample-level tokenization effectively compresses MEG data into a compact discrete space while preserving temporal and spatial resolution. The strong cross-dataset generalization further suggests that these tokenizers capture universal, data-invariant features of MEG signals rather than fitting to dataset-specific spatiotemporal or spectral patterns, leaving such structure to be modeled by the subsequent foundation model.

\subsection{Token Prediction Analysis}

Although tokenizers can be evaluated independently, their practical utility ultimately depends on how effectively they support foundation model training. To assess this, we trained MEG-GPT models using each tokenizer and evaluated token prediction performance on the Cam-CAN validation set. Given an input token sequence, the pretrained MEG-GPT was tasked with predicting the next token in an autoregressive manner. Because each tokenizer induces a distinct discrete label space, predicted tokens were first de-tokenized into continuous-valued time points and compared against the corresponding ground-truth signals. Prediction accuracy was quantified by computing channel-wise PVE over the time dimension from results across all validation sequences.

\begin{figure}[!t]
\centerline{\includegraphics[width=\columnwidth]{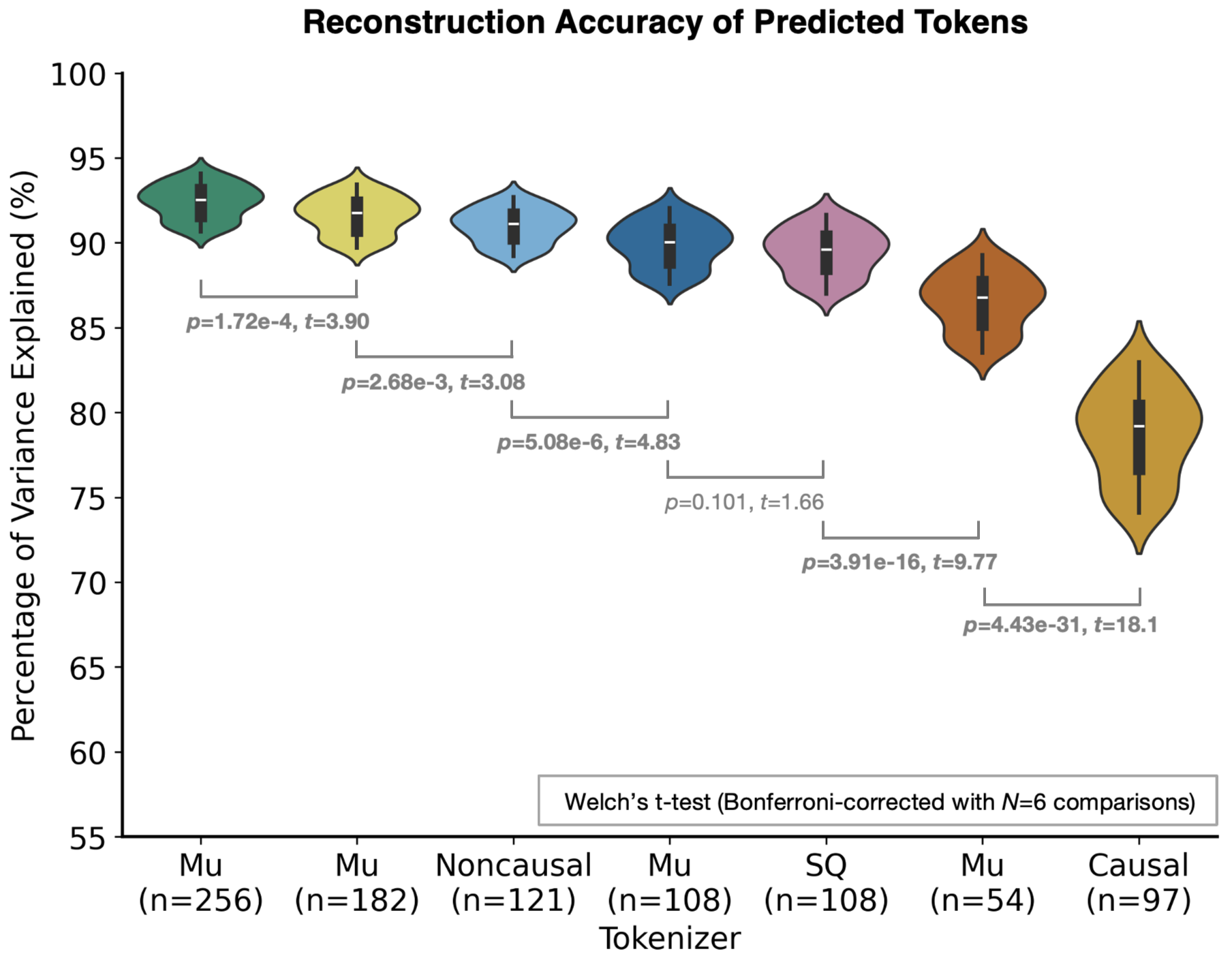}}
\caption{Token prediction accuracy of MEG-GPT pretrained with different tokenizers, evaluated on the Cam-CAN validation set. Tokenizers are ordered by descending accuracy. Statistical significance between adjacent pairs was assessed using Welch's t-test (after verifying normality and equal variance assumptions), with Bonferroni correction for multiple comparisons; significant differences are shown in bold.}
\label{fig6}
\end{figure}

As shown in \hyperref[fig6]{Fig.~6}, $\mu$-transform tokenizers with larger vocabularies achieved significantly higher reconstruction accuracy than other tokenization schemes. Overall performance trends were broadly consistent with the reconstruction results in \hyperref[fig5]{Fig.~5}, with the exception of the noncausal tokenizer. This deviation is partly attributable to the inherent design of a noncausal tokenizer, whose token representations incorporate information from both past and future time points. When paired with a causal AR foundation model, its convolution kernels may have introduced temporal information leakage that artificially benefits token prediction performance. While care was taken to ensure a fair comparison across tokenizers, this effect should be considered when interpreting the results. Additionally, differences in outlier handling during the de-tokenization process between learnable and non-learnable tokenizers may further contribute to observed performance discrepancies.

\begin{figure*}[!t]
\centerline{\includegraphics[width=7.16in]{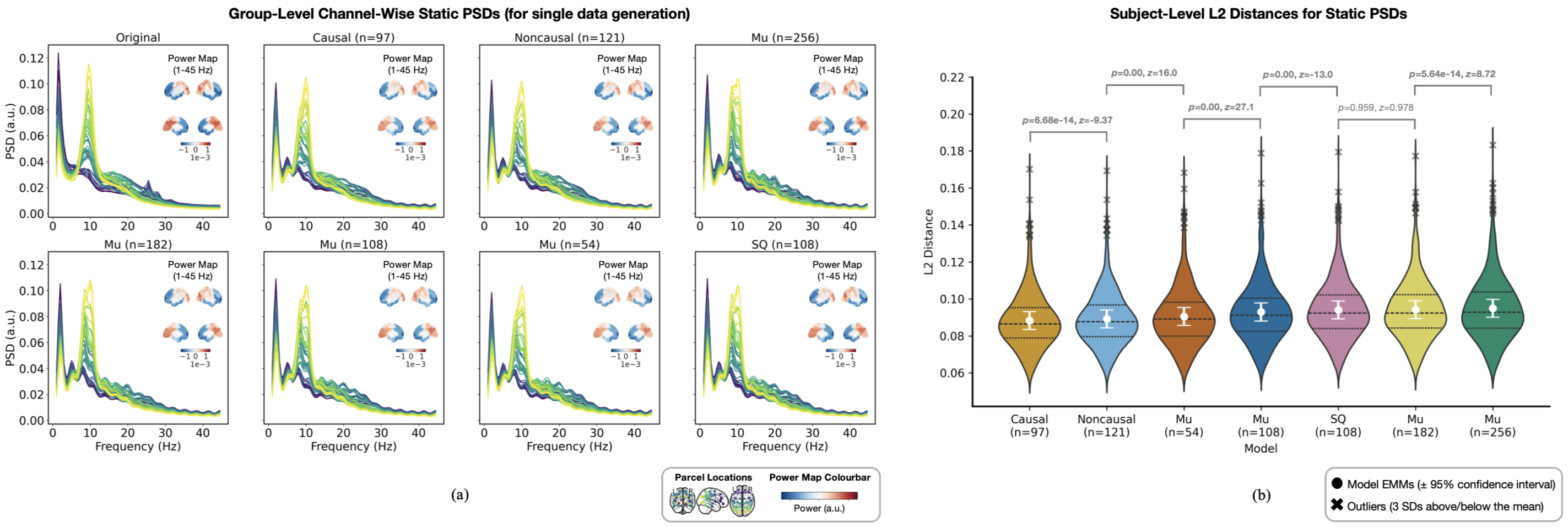}}
\caption{Static spectral characteristics of data generated by MEG-GPT models trained with different tokenizers. (a) Group-level channel-wise PSDs from a single data generation instance; insets show spatial power maps averaged over 1-45~Hz. (b) Subject-level L2 distances between static PSDs of generated and real data, averaged over channels and datasets. Statistical significance between adjacent pairs was assessed using Welch’s t-test with Bonferroni correction; significant differences are indicated in bold.}
\label{fig7}
\end{figure*}

\begin{figure*}[!t]
\centerline{\includegraphics[width=7.16in]{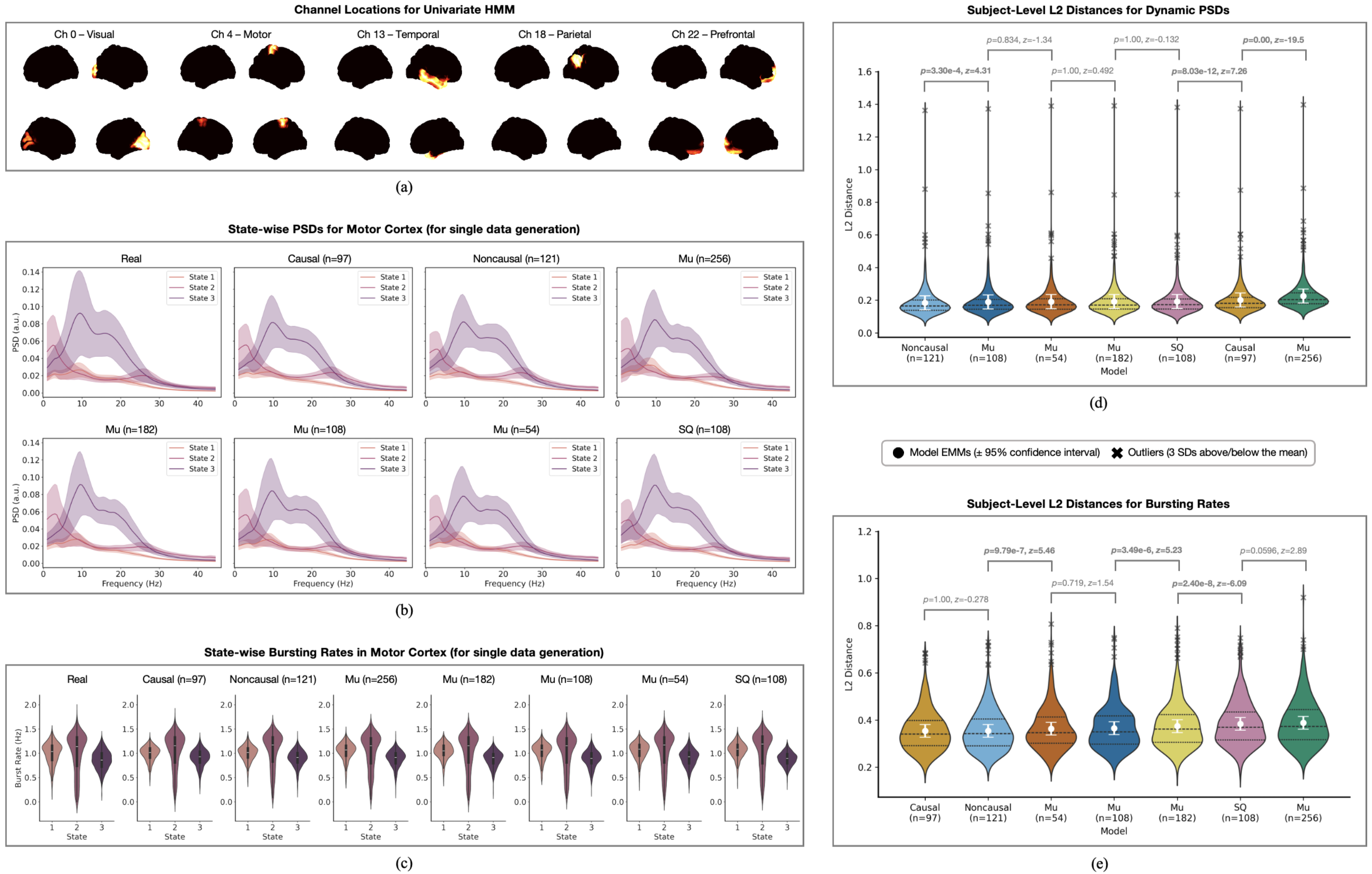}}
\caption{Dynamic spectral bursting characteristics of data generated by MEG-GPT models trained with different tokenizers. (a) Brain regions selected for bursting analysis. A three-state univariate TDE-HMM was applied to each region. (b) State-wise group-level PSDs for the motor cortex from a single data generation instance. (c) State-wise subject-level bursting rates for the motor cortex from the same data generation. (d) Subject-level L2 distances between generated and real state-wise PSDs, averaged over selected channels and datasets. (e) Subject-level L2 distances between generated and real state-wise bursting rates, averaged over selected channels and datasets. Statistical significance between adjacent pairs was assessed using Welch’s t-test with Bonferroni correction; significant differences are indicated in bold.}
\label{fig8}
\end{figure*}

\subsection{Generated Data Quality}

Next, we evaluated the impact of tokenizer choice on the quality of MEG-GPT-generated data by analyzing the spatial and spectral properties of synthetic signals, focusing on both static spectral structure and dynamic bursting behavior.

\subsubsection{Static Spectral Analysis}

Synthetic MEG data were generated using MEG-GPT models trained with each tokenizer, and static power spectral densities (PSDs) were estimated using Welch's method (2~s windows, 50\% overlap). \hyperref[fig7]{Fig.~7(a)} shows group-level (subject-averaged) channel-wise PSDs from a single data generation per tokenizer, along with spatial power maps obtained by averaging PSDs over 1–45~Hz. All models reproduced characteristic spectral profiles of real MEG data, including prominent delta and alpha peaks, and yielded comparable spatial power distributions with maximal activation in parietal and visual cortices. Notably, non-learnable baseline tokenizers exhibited noticeable ripple artifacts in higher-frequency bands ($>$10~Hz).

For quantitative comparison, subject-level PSDs of size $(N, C, F)$—where $N$ denotes subjects, $C$ channels, and $F$ frequency bins—were computed for both real and generated data. L2 distances between real and synthetic PSDs were calculated along the frequency dimension, yielding subject–channel–level error measures. These distances were analyzed using a linear mixed-effects model with fixed effects for tokenizer, dataset, and their interaction, and random intercepts for subject and channel. Fixed effects were assessed via Type~III ANOVA with Satterthwaite degrees-of-freedom approximation, followed by estimation of marginal means (EMMs) and Tukey-corrected pairwise comparisons. As depicted in \hyperref[fig7]{Fig.~7(b)}, learnable tokenizers achieved significantly lower L2 distances than non-learnable baselines, indicating improved preservation of static spectral structure.

\subsubsection{Dynamic Spectral Analysis}

We next investigated whether MEG-GPT captures transient spectral bursting, a hallmark of neural dynamics not observable under temporal or trial averaging \cite{Quinn_2019}. Bursting analysis was performed in five distinct brain regions (\hyperref[fig8]{Fig.~8(a)}). For each region, corresponding parcel time courses were extracted and modeled using a three-state univariate time-delay embedded hidden Markov model (TDE-HMM) to characterize state-specific oscillatory bursts \cite{Vidaurre_2018}. This procedure was repeated across ten independent data generations for each tokenizer.

\hyperref[fig8]{Figs.~8(b)–(c)} illustrate qualitative results from the motor cortex, a region known to exhibit prominent beta-band bursting \cite{Bonaiuto_2021}. Ground-truth states were established by fitting the same HMM to real data, which revealed states associated with delta/theta (State 2) and alpha/beta (State 3) activity. All tokenizers recovered three states with spectral profiles closely matching those of the real data, with pronounced beta bursting power. The temporal statistics of the states, quantified by bursting rates (i.e., average bursts per second), were also comparable across tokenizers and consistent with ground truth.

Dynamic similarity between generated and real data was quantified by computing L2 distances between generated and real state-wise PSDs (across frequency and state dimensions) and between state-wise bursting rates (across states). Statistical comparisons followed the same mixed-effects framework as in the static analysis. Noncausal tokenizers achieved lower L2 distances for dynamic PSDs (\hyperref[fig8]{Fig. 8(d)}), while both learnable tokenizers outperformed baseline methods in reproducing bursting rates (\hyperref[fig8]{Fig. 8(e)}).

In summary, MEG-GPT models generated data that faithfully captured both static and dynamic spatio-spectral characteristics of real MEG signals across all tokenizer types. Although some tokenizer-dependent differences reached statistical significance, their effect sizes were small, likely reflecting the large sample size and limiting strong interpretive claims. Within this context, while learnable tokenizers generally exhibited improved qualitative and quantitative advantages (including reduced spectral ripples), the results indicate that all tokenizer types achieve broadly comparable generative performance in modeling MEG spectral structure.

\subsection{Subject Fingerprinting and Inter-Subject Variability}

When pre-training MEG-GPT, subject labels were provided to learn subject-specific embeddings (\hyperref[fig2]{Fig.~2(a)}). We therefore examined how MEG-GPT models trained with different tokenizers differ in their ability to capture subject-specific neural signatures and generate data that preserve inter-subject variability.

Subject fingerprinting was evaluated by extracting subject-level features from both real and generated data and performing subject identification using a nearest-neighbor classifier with correlation-based distance. Features were derived by applying a TDE transformation to the data \cite{Vidaurre_2016, Huang_2025}, augmenting each channel with its lagged copies to capture joint spatio-spectral information. For each subject, a static covariance matrix was computed from the TDE data, and the upper triangular elements were vectorized to produce the features.

\begin{figure}[!t]
\centerline{\includegraphics[width=\columnwidth]{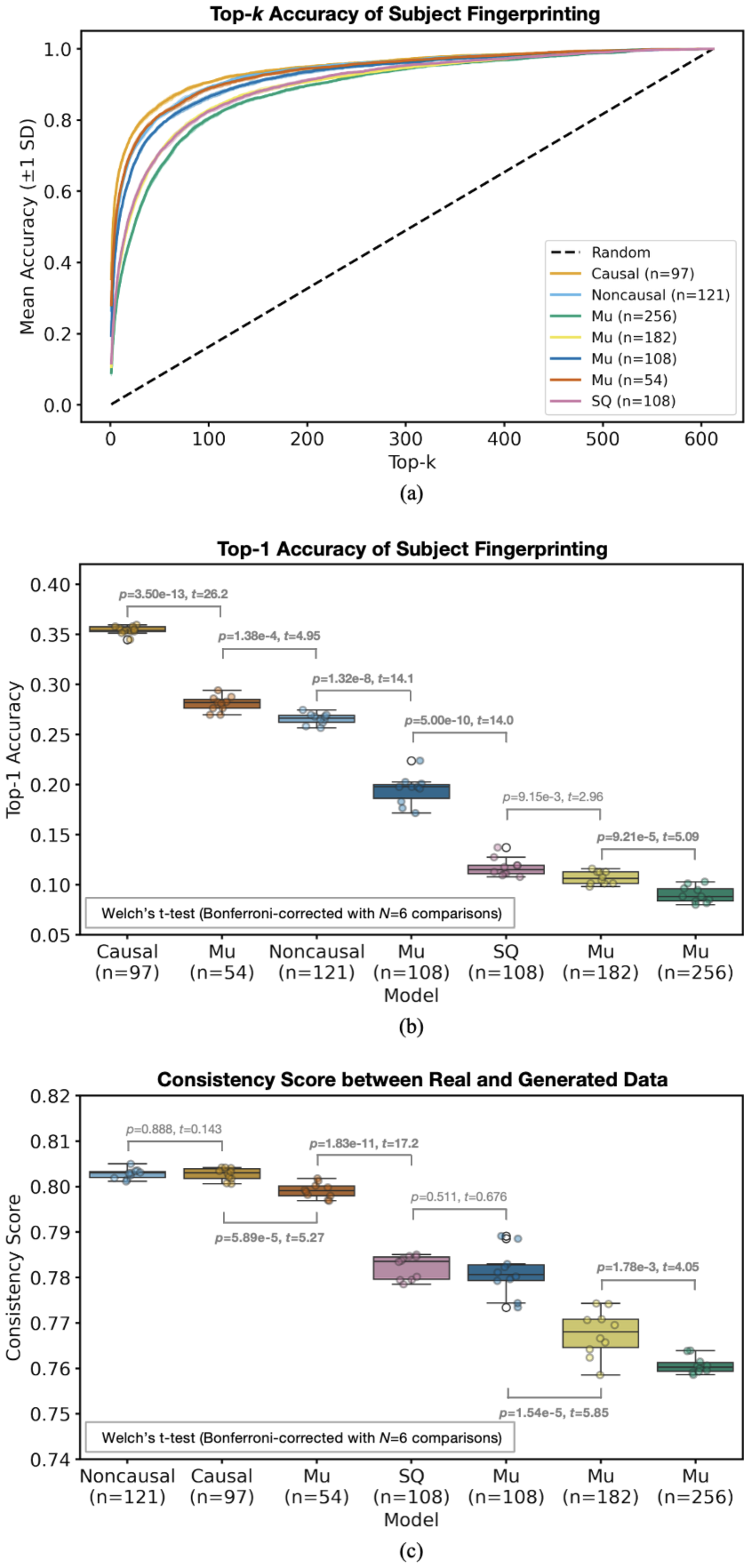}}
\caption{Subject fingerprinting and inter-subject similarity in MEG-GPT-generated data. (a) Top-$k$ subject prediction accuracy based on TDE features, shown as mean $\pm$ 1 standard deviation across ten data generation instances. (b) Top-1 accuracy across ten data generations, ordered by mean accuracy. (c) Consistency scores across ten data generations, ordered by mean score. Statistical significance between adjacent pairs was assessed using Welch’s t-test with Bonferroni correction; significant differences are indicated in bold.}
\label{fig9}
\end{figure}

We first quantified the accuracy of predicting subject labels from MEG-GPT–generated data based on these TDE features. Let $f_i$ and $f'_j$ denote features from subject $i$ in the real data $x$ and subject $j$ in the generated data $x'$, respectively. A pairwise distance matrix $\Sigma_{x,x'} \in \mathbb{R}^{N \times N}$ was constructed with entries given by $1-\mathrm{corr}(f_i,f'_j)$. Top-$k$ accuracy was defined as the proportion of subjects for which the diagonal entry was among the $k$ smallest distances in the corresponding column. As shown in \hyperref[fig9]{Fig.~9(a)}, tokenizer-dependent differences persisted up to $k \approx 300$. Focusing on top-1 accuracy (\hyperref[fig9]{Fig.~9(b)}), the causal tokenizer achieved the highest performance with a statistically significant and large effect size, followed by the $\mu$-transform tokenizer with 54 tokens and the noncausal tokenizer.

To additionally investigate how well MEG-GPT preserves inter-subject relationships across tokenizers, we computed a consistency score measuring agreement between inter-subject similarity structures in real and generated data. Specifically, correlation matrices $\Sigma_{x,x}$ and $\Sigma_{x',x'}$ were computed across subjects for real and generated features, respectively, and the consistency score was defined as the Pearson correlation between their upper triangular elements. This metric quantifies the extent to which inter-subject relationships in the original data are preserved. As shown in \hyperref[fig9]{Fig.~9(c)}, learnable tokenizers achieved significantly higher consistency scores with large effect sizes compared to non-learnable baselines.

In summary, learnable tokenizers enabled MEG-GPT to more effectively capture subject-specific spatio-spectral signatures and inter-subject variability. In contrast, baseline tokenizers—particularly those with larger vocabulary size—exhibited reduced capacity to preserve these subject-level characteristics.

\subsection{Downstream Decoding Task Performance}

\begin{figure}[!t]
\centerline{\includegraphics[width=\columnwidth]{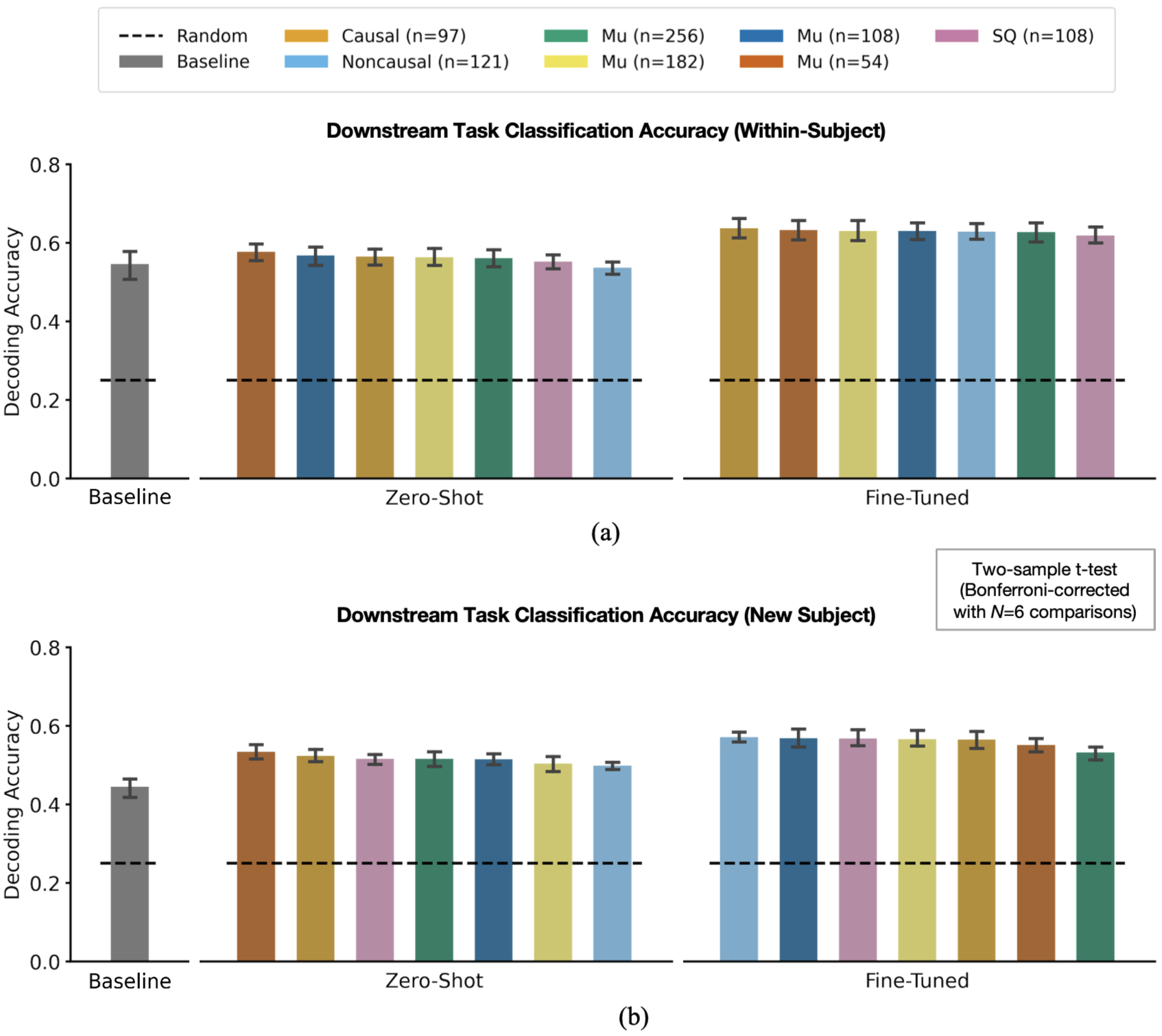}}
\caption{Downstream task decoding performance of MEG-GPT models trained with different tokenizers. Task classification accuracy is shown for (a) within-subject and (b) new-subject conditions under baseline, zero-shot, and fine-tuned regimes. Statistical significance between adjacent pairs was assessed using two-sample t-tests with Bonferroni correction; no significant differences were observed.}
\label{fig10}
\end{figure}

Finally, we evaluated the performance of MEG-GPT models on a downstream task decoding setting. For each tokenizer, a single group-level multinomial logistic regression classifier was trained to predict four task labels (see Section~IV-C). Performance was evaluated under two conditions: \emph{within-subject} (\hyperref[fig10]{Fig.~10(a)}), where task labels were predicted for unseen sessions from subjects included during training, and \emph{new-subject} (\hyperref[fig10]{Fig.~10(b)}), where task labels were predicted for an entirely held-out participant. Prediction accuracy was calculated for the baseline, zero-shot, and end-to-end fine-tuning regimes across all tokenizers.

Across both conditions, zero-shot decoding consistently outperformed the baseline, and end-to-end fine-tuning further improved performance for all tokenizer types. These results indicate that MEG-GPT learns transferable representations of neural activity that enhance the performance of simple linear classifiers and can be effectively adapted for downstream decoding tasks. In the fine-tuned regime, causal and noncausal tokenizers achieved the highest classification accuracy in the within-subject and new-subject settings, respectively. However, differences in decoding accuracy between tokenizers were not statistically significant in either the zero-shot or fine-tuned regimes, suggesting that all tokenizers exhibit comparable downstream classification capacity.

\section{Discussion}

In this work, we systematically evaluated sample-level tokenizers for LNMs applied to MEG data across multiple criteria, including reconstruction fidelity, generative data quality, subject fingerprinting, and downstream decoding performance. Across all these dimensions, both learnable and non-learnable tokenizers exhibited broadly comparable performance, suggesting that learnable tokenization provides limited empirical benefit for sample-level neural representations. This supports a practical design principle whereby LNMs for neural time series can rely on simple, fixed discretization schemes without substantial performance degradation, particularly when computational efficiency and scalability are primary considerations.

Nonetheless, the advantages observed for learnable tokenizers should not be dismissed. Although $\mu$-transform tokenizers achieved higher token prediction accuracy (Fig. 6), learnable tokenizers demonstrated modest but consistent improvements in capturing spatio-spectral structure in generated data (Figs. 7, 8), as well as in yielding improved performance under end-to-end fine-tuning for downstream decoding (Fig. 10). Learnable tokenizers also more effectively preserved subject-specific neural fingerprints and inter-subject similarity structure (Fig. 9). Together, these results indicate that learnable tokenization may be preferable in applications where subject individuality or fine-grained generative fidelity is a primary objective.

A key aspect of our evaluation framework is that it extends beyond downstream decoding accuracy. While task performance is an important indicator of representation utility, it provides only a partial view of model behavior, especially for biologically grounded neural data. Our results highlight the importance of evaluating LNMs based on their ability to generate realistic synthetic signals that capture non-stationary dynamics, transient spectral events, subject-specific structure, and biologically meaningful spatio-spectral characteristics.

Our analysis also highlights constraints inherent to both learnable and non-learnable tokenizers. Non-learnable tokenizers rely on data-driven scaling and binning procedures that may not generalize across datasets with different sample distributions, potentially requiring dataset-specific recalibration. Similar considerations apply to learnable tokenizers, whose parameters are likewise data-adaptive. However, as shown in Fig.~5(b), this adaptivity did not impair generalization across the MEG datasets examined here, which span different acquisition sites, hardware, and experimental paradigms. Whether such robustness extends to non-sample-level tokenizers—where temporal or spatial context is explicitly compressed into tokens—remains an open question.

Another important consideration is the identifiability of the learnable tokenizers. As these tokenizers are trained using a standard autoencoder, the inferred token representations are not uniquely identifiable. In particular, multiple distinct sets of token assignments may reconstruct the data equally well, leading to variability in the inferred tokens even when trained on the same dataset. This non-uniqueness arises from the fundamental indeterminacy of latent representations in autoencoders: given an encoder $f$ and a decoder $g$, the reconstruction $g(f(x))$ remains invariant under arbitrary invertible transformations $G$ of the latent space, i.e., $g(f(x)) = g(G^{-1}(G(f(x))))$. Consequently, unless additional priors or inductive biases are imposed, there may exist infinitely many equivalent latent representations that explain the observed data. This property does not affect reconstruction accuracy but may complicate the interpretability and comparability of learned tokens across training runs.

Finally, this study has several limitations. First, due to limited computational resources, MEG-GPT models were pretrained and fine-tuned only once per tokenizer, leaving run-to-run variability uncharacterized. Although stochasticity was controlled through fixed random seeds to ensure fair comparison, repeated training runs would enable more robust statistical assessment. Second, the receptive field of MEG-GPT was limited to 80 samples (320 ms), which may restrict sensitivity to slower neural dynamics and bias evaluations of generative quality, particularly for learnable tokenizers trained with longer temporal contexts. Third, our analyses relied on source-reconstructed parcel-level time courses with fixed preprocessing choices, including sampling rate and bandpass filtering. The impact of alternative preprocessing pipelines on tokenization and model performance warrants further investigation.

Future work may extend this framework in two primary directions. First, comparisons could be broadened to include non-sample-level tokenizers and alternative foundation models for neural time series, enabling a more comprehensive assessment of trade-offs between tokenization strategies. Second, extending evaluations across modalities such as EEG, optically pumped magnetometer (OPM)-MEG, and functional magnetic resonance imaging (fMRI) would strengthen our conclusions and clarify modality-specific considerations for neural time series tokenization in LNMs.

\section{Conclusion}

This paper presents the first systematic comparison of sample-level tokenization strategies for transformer-based neuroimaging foundation models using MEG data. By evaluating learnable and non-learnable tokenizers within a controlled experimental setting, we examined their effects on signal reconstruction, generative behavior, subject-level representation, and downstream decoding tasks. Our results show that simple fixed discretization schemes achieve performance comparable to learnable tokenizers across most evaluation criteria, even though learnable tokenizers provide modest gains in preserving subject-specific information and spatio-spectral structure in the model-generated data. These findings establish sample-level tokenization as a viable and effective design choice for neural foundation models and offer empirical guidance for selecting tokenization strategies based on modeling objectives and practical considerations.

\end{document}